\documentclass[11pt]{article} % use larger type; default would be 10pt

\usepackage[utf8]{inputenc} % set input encoding (not needed with XeLaTeX)

\usepackage{geometry} % to change the page dimensions
\usepackage{graphicx} % support the \includegraphics command and options

\usepackage[singlelinecheck=false]{caption}

\usepackage{booktabs} % for much better looking tables
\usepackage{array} % for better arrays (eg matrices) in maths
\usepackage{paralist} % very flexible & customisable lists (eg. enumerate/itemize, etc.)
\usepackage{verbatim} % adds environment for commenting out blocks of text & for better verbatim
\usepackage{subfig} % make it possible to include more than one captioned figure/table in a single float
\usepackage{amsthm,amsmath}
\usepackage{float}
\usepackage{setspace}

\usepackage{dsfont} % indicator function

\usepackage{fancyhdr} % This should be set AFTER setting up the page geometry
\usepackage{sectsty}
\allsectionsfont{\sffamily\mdseries\upshape} % (See the fntguide.pdf for font help)
\usepackage[nottoc,notlof,notlot]{tocbibind} % Put the bibliography in the ToC
\usepackage[titles,subfigure]{tocloft} % Alter the style of the Table of Contents

\usepackage{array}
\usepackage{multirow}
\newcolumntype{L}[1]{>{\raggedright\let\newline\\\arraybackslash\hspace{0pt}}m{#1}}
\newcolumntype{C}[1]{>{\centering\let\newline\\\arraybackslash\hspace{0pt}}m{#1}}
\newcolumntype{R}[1]{>{\raggedleft\let\newline\\\arraybackslash\hspace{0pt}}m{#1}}

\usepackage{xcolor}

\definecolor{yellow}{RGB}{208, 218, 56}
\definecolor{green}{RGB}{120, 191, 61}
\definecolor{orange}{rgb}{0.99,0.69,0.07}

\usepackage{pifont} % ding
\usepackage{authblk}

\title{Individual dynamic prediction of clinical endpoint from large dimensional longitudinal biomarker history: a landmark approach}
\author[1,*]{Anthony Devaux}
\author[1,2]{Robin Genuer}
\author[1]{Karine Pérès}
\author[1]{Cécile Proust-Lima}

\affil[1]{INSERM, BPH, U1219, University of Bordeaux, Bordeaux, France}
\affil[2]{INRIA Bordeaux Sud-Ouest, Talence, France}
\affil[*]{Email: anthony.devaux@u-bordeaux.fr}
\date{January 2022} % Activate to display a given date or no date (if empty),
\begin{document}

\singlespacing

\maketitle

\textbf{Abstract:} The individual data collected throughout patient follow-up constitute crucial information for assessing the risk of a clinical event, and eventually for adapting a therapeutic strategy. Joint models and landmark models have been proposed to compute individual dynamic predictions from repeated measures to one or two markers. However, they hardly extend to the case where the patient history includes much more repeated markers. Our objective was thus to propose a solution for the dynamic prediction of a health event that may exploit repeated measures of a possibly large number of markers. We combined a landmark approach extended to endogenous markers history with machine learning methods adapted to survival data. Each marker trajectory is modeled using the information collected up to the landmark time, and summary variables that best capture the individual trajectories are derived. These summaries and additional covariates are then included in different prediction methods adapted to survival data, namely regularized regressions and random survival forests, to predict the event from the landmark time. We also show how predictive tools can be combined into a superlearner. The performances are evaluated by cross-validation using estimators of Brier Score and the area under the Receiver Operating Characteristic curve adapted to censored data. We demonstrate in a simulation study the benefits of machine learning survival methods over standard survival models, especially in the case of numerous and/or nonlinear relationships between the predictors and the event. We then applied the methodology in two prediction contexts: a clinical context with the prediction of death in primary biliary cholangitis, and a public health context with age-specific prediction of death in the general elderly population. Our methodology, implemented in R, enables the prediction of an event using the entire longitudinal patient history, even when the number of repeated markers is large. Although introduced with mixed models for the repeated markers and methods for a single right censored time-to-event, the technique can be used with any other appropriate modeling technique for the markers and can be easily extended to competing risks setting. \\


%%%%%%%%%%%%%%%%
%% Background %%
%%
\section{Background}

\begin{figure}[H]
  \includegraphics{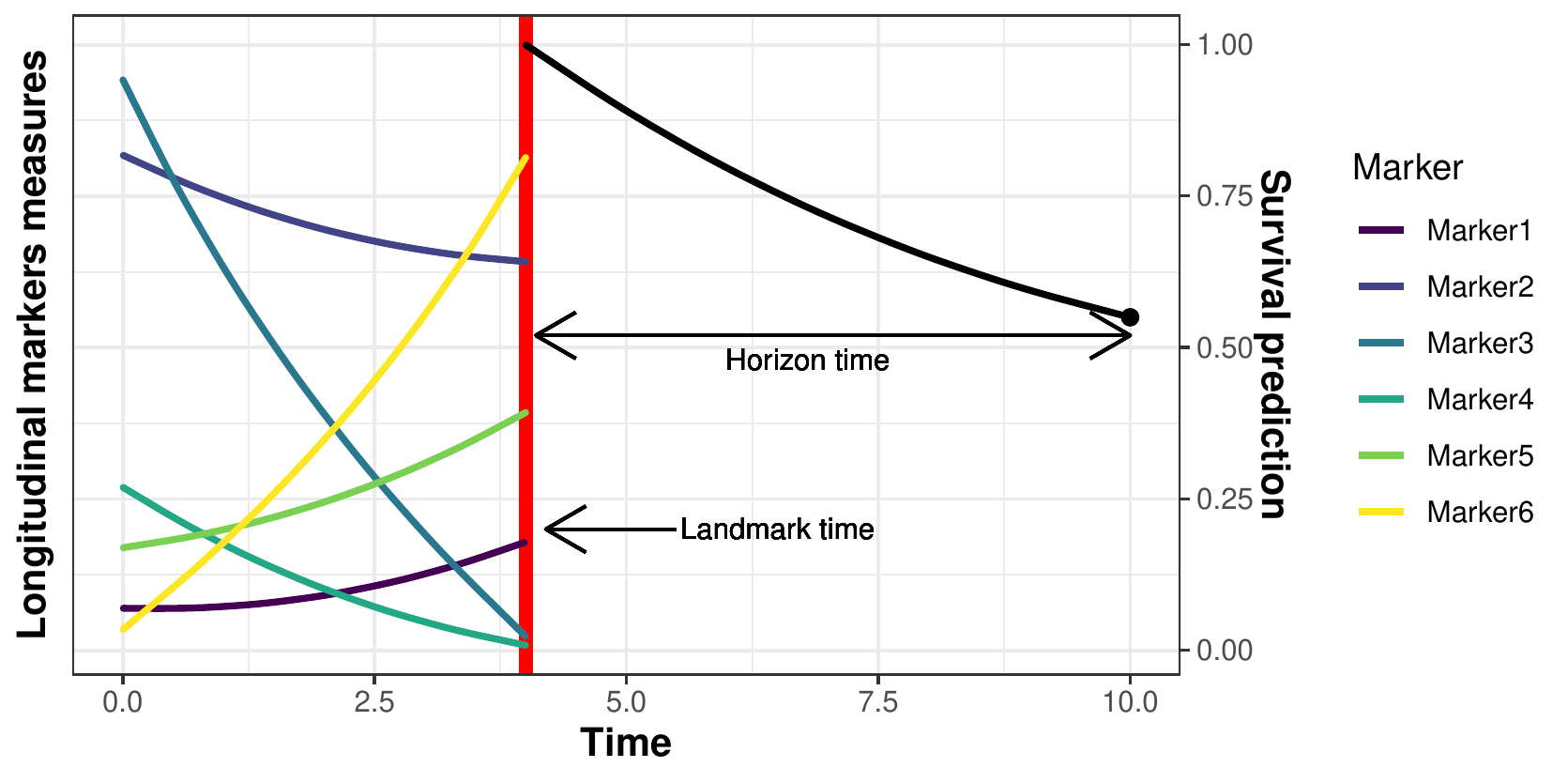}
  \caption{Illustration of individual dynamic prediction of an event computed using history of multiple repeated markers (here 6). The individual probability of event is computed from a landmark time to a horizon time by using the information on the markers trajectories collected up to the landmark time.}
  \label{fig:illus_hdlandmark}
\end{figure}

A central issue in health care is to quantify the risk of disease, disease progression or death at the individual level, for instance to initiate or adapt a treatment strategy as soon as possible. To achieve this goal, the information collected at a given time (at diagnosis or at the first visit) is often not sufficient and repeated measurements of markers are essential. For example, repeated prostate specific antigen (PSA) data are highly predictive of the risk of prostate cancer recurrence \cite{proust-lima_development_2009,sene_shared_2014,taylor_real-time_2013}, and markers such as diabetic status or blood pressure level over time are crucial in predicting the risk of cardiovascular disease \cite{paige_landmark_2018,sweeting_use_2017}. Including longitudinal information into the prediction of a clinical event defines the framework for individual dynamic predictions \cite{proust-lima_development_2009,rizopoulos_dynamic_2011,ferrer_individual_2019}. In some contexts, a single marker may be sufficient to predict the occurrence of the event (e.g., in prostate cancer with PSA) but often the complete patient history with possibly many repeated markers should be exploited (see Figure \ref{fig:illus_hdlandmark}). Yet, statistical developments for individual prediction of event have so far either focused on the repeated nature of the information or on its large dimension. 

When using repeated information to develop dynamic prediction tools, two approaches are commonly used: joint models \cite{proust-lima_development_2009,rizopoulos_dynamic_2011} and landmark models \cite{van_houwelingen_dynamic_2007}. Joint models simultaneously analyze the longitudinal and event time processes by assuming a structure of association built on summary variables of the marker dynamics \cite{tsiatis_joint_2004}. This model which uses all the information on the longitudinal and time-to-event processes to derive the prediction tool is widely used in the case of a single repeated marker but becomes intractable in the presence of more than a few repeated markers due to high computational complexity \cite{ferrer_individual_2019}. 

An alternative is to use partly conditional survival model \cite{maziarz_longitudinal_2017} or landmark models \cite{van_houwelingen_dynamic_2007} which consist in directly focusing on the individuals still at risk at the landmark time and consider their history up to the landmark time (see Figure \ref{fig:illus_hdlandmark}). When individual history includes repeated measures of an endogenous marker, summaries of the marker derived from preliminary mixed models can be included in the survival model, instead of only the last observed value \cite{proust-lima_development_2009, sweeting_use_2017}. Although the landmark models do not use as much information as the joint model (only information from the at-risk individuals at the landmark time is exploited) and thus may lack of efficiency, they have shown competitive predictive performances, easier implementation (much less numerical problems) and better robustness to misspecification than joint models \cite{ferrer_individual_2019}. However, as joint models, they necessitate to consider the actual nature of the relationship between the marker and the event. 

Although the landmark approach is \textit{per se} very general, in practice its definition is based on standard survival models, namely the Cox model, which prevents the methodology to be applied in large dimensional contexts usually encountered in applications. Indeed the Cox model becomes rapidly limited in the presence of: 1) a large number of predictors, 2) highly correlated predictors, and 3) complex relationships between the predictors and the event \cite{goldstein_moving_2016}. Yet, in the context of dynamic prediction from multiple repeated markers, these three limits are rapidly reached. Indeed, the large dimension of the predictors does not only come from the number of markers but also from the number of (potentially correlated with each other) marker-specific summaries that are necessary to approximate the actual nature of the relationship between the marker and the event. 

Machine learning methods, including regularized regressions or decision trees and random forests, have been specifically developed to predict outcomes while tackling the aforementioned issues \cite{breiman_random_2001}. Their good predictive performances have been largely demonstrated in the literature \cite{lebedev_random_2014}. Initially proposed for continuous or binary outcomes, they have been recently extended to handle right censored time-to-event data. For instance, Simon \textit{et al.}  \cite{simon_regularization_2011} developed penalized Cox models either using Ridge, Lasso or Elastic-Net penalty, Bastien \textit{et al.} \cite{bastien_deviance_2015} developed a Cox model based on deviance residuals-based sparse-Partial Least Square, as an extension of sparse-Partial Least Square \cite{chun_sparse_2010} for survival data, and Ishwaran \textit{et al.} \cite{ishwaran_random_2008} extended random forests to survival data. However, they were mostly applied to predict time-to-event from time-independent marker information. Our purpose is thus to show how these machine learning methods can also be leveraged to provide dynamic individual predictions from large dimensional longitudinal biomarker data. 

Computing dynamic predictions in the context of a large number of repeated markers is a very new topic in statistics, and only a few proposals have been made very recently. Zhao \textit{et al.} \cite{zhao_incorporating_2020} and Jiang \textit{et al.} \cite{jiang_functional_2020} focused on random forests. Using a landmark approach, Zhao \textit{et al.} transformed the survival data into pseudo-observations and incorporated in each tree the marker information at a randomly selected time. Although handling repeated markers, this method neither accounts for measurement errors of the biomarkers nor their trajectory shapes. By considering a functional ensemble survival tree, Jiang \textit{et al.} overcame this issue. They characterized the changing patterns of continuous time-varying biomarkers using functional data analysis, and incorporated those characteristics directly into random survival forests. By concomitantly analyzing the markers and the event, this approach belongs to the two-stage calibration approaches \cite{ye_semiparametric_2008} and may suffer from the same biases \cite{albert_estimating_2010}. Finally Tanner \textit{et al.} \cite{tanner_dynamic_2020} proposed to extend the landmark approach to incorporate multiple repeated markers with measurements errors. For the survival prediction method, they chose to discretize the time and use an ensemble of classical binary classifiers to predict the event.

In comparison with this emerging literature, our proposal goes one step forward. As in Tanner \textit{et al.}, we chose to rely on a landmark approach and consider various prediction methods rather than only random forests. However, we also chose to directly exploit the survival data in continuous time. In addition, our methodology handles markers of different nature, accounts for their measurement error and intermittent missing data, and for a possibly large number of summary characteristics of each marker. 

In the following sections, we first describe the proposed method. We then demonstrate in a simulation study the performances of the methodology and the benefit of using machine learning methods to handle the large dimensional aspect. We then illustrate the methodology in two very different contexts: a clinical context with the prediction of death in primary biliary cholangitis, and a public health context with the prediction of 5-year death at different ages in the general elderly population. The paper ends with the discussion of the strengths and weaknesses of the proposed method.

%%%%%%%%%%%%%
%% Methods %%
%%
\section{Methods}

\subsection{Framework, notations and general principle}

Let us consider a landmark time $t_{LM}$ of interest and a population of $N_{t_{LM}}$ individuals that are still at risk of the event at $t_{LM}$. For an individual $i \in \{1,\dots,N_{t_{LM}}\}$, we denote $T_i$ the true event time, $C_i$ the independent censoring time. We define $T_i^\star = \min{(T_i, C_i)}$ the observed time event and $\delta_i = \mathds{1}\left(T_i < \min{(C_i,t_{LM} + t_{Hor})} \right)$ the event indicator with $t_{Hor}$ the horizon time. We consider a single event for simplicity. 

At the landmark time, $P$ time-independent covariates $X_i$ are available, and the history of $K$ time-dependent markers $Y_{ijk}$ ($k \in \{1,\dots,K\}$) measured at time $t_{ij}$ ($j \in \{1,\dots,n_i\}$) and $t_{ij} \leq t_{LM}$. 

The target individual probability of event from the landmark time $t_{LM}$ to the horizon time $t_{Hor}$ of a subject $\star$ is defined as:
\begin{equation}
    \pi_{\star}(t_{LM},t_{Hor}) =  P (T_\star \leq t_{LM} + t_{Hor}~|~ T_\star > t_{LM}, \{Y_{{\star}jk}; k=1,...,K, t_{{\star}jk}\leq t_{LM} \}, X_{\star} ) \label{eq:proba_indiv}
\end{equation}
By assuming that the history of the $K$ marker trajectories up to $t_{LM}$ can be summarized into a vector $\Gamma_\star$, we define the following probability:
\begin{equation}
\widetilde{\pi}_{\star}(t_{LM},t_{Hor}) =  P(T_\star \leq t_{LM} + t_{Hor}~|~ T_\star > t_{LM}, \Gamma_\star(t_{LM}), X_\star) \label{eq:proba_indiv_estimator}
\end{equation}

This probability is estimated by $\widehat{\pi}_{\star}^{(m)}(t_{LM},t_{Hor})$ in 4 steps on a learning sample for each survival prediction method $m$:
\begin{enumerate}
\item Each marker trajectory is modeled using the information collected up to $t_{LM}$
\item The vector of summary variables $\Gamma_i(t_{LM})$ is computed for each individual $i$
\item $\Gamma_i(t_{LM})$ and additional baseline covariates $X_i$ are entered into survival prediction method $m$
\item The predicted probability of event $\widehat{\pi}_{\star}^{(m)}(t_{LM},t_{Hor})$ is computed from survival method $m$
\end{enumerate}

Once the estimator defined (i.e., the survival prediction method trained) on the learning sample, the summary variables $\Gamma_{\star}(t_{LM})$ can be computed for any new external individual $\star$ at risk of event at $t_{LM}$, and the corresponding individual predicted probability of event can be deduced.

\subsection{Step 1. Longitudinal model for markers history}

Longitudinal markers are usually measured at intermittent times with error. The first step consists to estimate the error-free trajectory of the marker of each individual over the history period. We propose to use generalized mixed models \cite{laird_random-effects_1982} defined as:
\begin{eqnarray}
\begin{split}
g(E(Y_{ijk}|b_{ik})) & = Y_{ik}^\ast(t_{ijk}) = X^{\top}_{ik}(t_{ijk}) \beta_k + Z^{\top}_{ik}(t_{ijk}) b_{ik} \label{GLMM}
\end{split}
\end{eqnarray}
\noindent where $X^{\top}_{ik}(t_{ijk})$ and $Z^{\top}_{ik}(t_{ijk})$ are the $p_k$- and $q_k$-vectors associated with the fixed effects $\beta_k$ and random effects $b_{ik}$ (with $b_{ik} \sim \mathcal{N}(0,B_k)$), respectively. The link function $g(.)$ is chosen according to the nature of $Y_{ijk}$ (e.g. identity function for Gaussian continuous markers or logit function for binary markers).

\subsection{Step 2. Summary characteristics of the marker trajectories} \label{sec:resum}

Once the parameters of the model have been estimated (indicated by $\widehat{\; \; }$ below), any summary that captures the marker behavior up to the time $t_{LM}$ can be computed. We give here a non-exhaustive list for individual $i$:
\begin{itemize}
\item Predicted individual deviations to the mean trajectory: $\widehat{b}_{ik} = \widehat{B}_k Z^{\top}_{ik} \widehat{V}^{-1}_{ik} (Y_{ik} - X_{ik} \widehat{\beta}_k)$ where $\widehat{V}_{ik} = Z_{ik} \widehat{B}_k Z^{\top}_{ik} + \widehat{\sigma}_{\epsilon k} I_{n_i}$, if the marker is continuous. Otherwise, $\widehat{b}_{ik} = \underset{b_{ik}}{\operatorname{argmax}}~f(b_{ik} | Y_{ik}^\ast) = \underset{b_{ik}}{\operatorname{argmax}}~f(Y_{ik}^\ast | b_{ik}) f(b_{ik})$ with $f(.)$ the density function;

\item Error-free level at time $u \leq t_{LM}$: $\widehat{Y}_{ik}^\ast(u) = X^{\top}_{ik}(u) \widehat{\beta}_k + Z^{\top}_{ik}(\tau) \widehat{b}_{ik}$ ;
\item Error-free slope at time $u \leq t_{LM}$: $\widehat{Y}^{\ast \prime}_{ik}(u) = \frac{\partial \widehat{Y}_{ik}^\ast(t)}{\partial t}|_{t=u}$ ;
\item Cumulative error-free level during period $\mathcal{T}$: $\widehat{h}_{ik}(t_{LM}) = \int_{t_{LM} - \mathcal{T}}^{t_{LM}} \widehat{Y}_{ik}^\ast(u) du$.
\end{itemize}

Any additional summary that is relevant for a specific disease can be considered as soon as it is a function of the error-free marker trajectory (e.g., time spent above/below a given threshold). All the individual summary characteristics across the $K$ markers are stored into a vector $\Gamma_i$. Using the list above and $u=t_{LM}$, $\Gamma_{i}(t_{LM}) = \{\Gamma_{ik}(t_{LM}), k=1,...,K \}$ with $\Gamma_{ik}(t_{LM})=(\widehat{b}_{ik}, \widehat{Y}_{ik}^\ast(t_{LM}), \widehat{Y}^{\ast \prime}_{ik}(t_{LM}), \widehat{h}_{ik}(t_{LM}))^\top$ is of length $\sum_{k=1}^K (q_k + 3)$. This vector may have a large amount of summaries which can also be highly correlated with each other. These particularities have to be taken into account in survival prediction methods.

\subsection{Step 3. Prediction methods for survival data in a large dimensional context}

To predict the risk of event from $t_{LM}$ to a horizon time $t_{Hor}$ using the vector $\mathcal{X}_i = (\Gamma_i,X_i)$ of summaries $\Gamma_i$ and time-independent variables $X_i$ of length $P$, we can use any technique that handles 1) right-censored time-to-event data, 2) the possibly high dimension, 3) and the correlation between the predictors. We focused in this work on Cox model, Penalized-Cox model, Deviance residuals-based sparse-Partial Least Square and Random Survival Forests, although other techniques could also be applied. For each technique, several sub-methods were considered that differ according to the type of variable selection and/or the hyperparameters choices. We briefly describe the different techniques and sub-methods below, and refer to Section 1 in supplementary material for further details.

\subsubsection{Cox models}

The Cox model is a semi-parametric regression which models the instantaneous risk of event according to a log-linear combination of the independent covariates:
\begin{equation}
\lambda_i(t|\Gamma_i,X_i) = \lambda_0(t) \exp{(X_i \gamma + \Gamma_i \eta)} \label{cox}
\end{equation}
with $\lambda_0$ the baseline hazard function, $\gamma$ and $\eta$ the coefficients estimated by partial likelihood. We defined two sub-models whether variable selection was performed according to backward selection procedure using \texttt{step}() R function (called \textit{Cox-SelectVar}) or not (\textit{Cox-AllVar}).

\subsubsection{Penalized-Cox models}

Penalized-Cox models extend the Cox model defined in \eqref{cox} to handle a high number of possibly correlated predictors. The partial log-likelihood is penalized with norm $\ell_2$ (Ridge penalty), norm $\ell_1$ (Lasso penalty \cite{goeman_l1_2009}) which enables covariate selection, or a mixture of both (Elastic-Net \cite{simon_regularization_2011}). These methods require the tuning of the norms mixing parameter (0 for Lasso, 1 for Ridge, $]0;1[$ for Elastic-Net) and the penalty parameter. We used \texttt{cv.glmnet}() function (from the \texttt{glmnet} R package) with internal cross-validation to tune the penalty parameter, and we defined three sub-models according to the norms mixing parameter (i.e. Lasso, Ridge or Elastic-Net). There are called \textit{Penal-Cox-Lasso}, \textit{Penal-Cox-Ridge} and \textit{Penal-Cox-Elastic}, respectively.

\subsubsection{Deviance residuals-based sparse-Partial Least Square (sPLS-DR)}

Partial Least Square (PLS) is a method of dimension reduction where components (or latent variables) are built to maximize the covariance with the outcome. Sparse-PLS (sPLS) \cite{chun_sparse_2010} adds a variable selection within each component using Lasso penalty. First developed in the framework of linear regression, this method was extended to survival data \cite{bastien_deviance_2015} (sPLS-DR). The principle is to apply a sPLS regression on the deviance residuals which are a normalized transformation of the martingale residuals $\widehat{\mathcal{M}}_i = \delta_i - \widehat{\Lambda}_i(t)$, with $\widehat{\Lambda}_i(t)$ the Nelson-Aalen cumulative hazard function estimate. Then, a Cox model is applied using the $C$ identified components $f_c(\Gamma_i, X_i)$ as covariates. In sPLS, the number of components $C$ and the Lasso penalty parameter on each component (which controls the sparsity on each component) have to be properly tuned. We used \texttt{cv.coxsplsDR}() function (from \texttt{plsRcox} R package) with internal cross-validation to tune the number of components, and considered three variants for the penalty: no penalty (called \textit{sPLS-NoSparse}), maximum penalty (called \textit{sPLS-MaxSparse}), or an optimized penalty from a grid of values (called \textit{sPLS-Optimize}).

\subsubsection{Random Survival Forests}

Random forests \cite{breiman_random_2001} are a non-parametric machine learning tool that can handle high-dimensional data with possibly complex input-output relationships. Random forests, originally developed in a context of regression or classification, were later adapted to right-censored survival data \cite{ishwaran_random_2008} and called random survival forests (RSF). A RSF aggregates $B$ survival trees, each one built on a different bootstrap sample from the original data (subjects not included in one bootstrap sample are called out-of-bag (OOB)). As any tree-based predictor, a survival tree recursively splits the sample into subgroups until the subgroups reach a certain minimal size $S$. To deal with time-to-event data, the splitting rule is usually based on the log-rank statistics although other splitting rules have also been proposed (e.g. gradient-based brier score splitting \cite{ishwaran_random_2008}). In RSF, at each node of each tree, a subset of $M$ predictors is randomly drawn and the split is optimized among splits candidates only involving those predictors. The size of the predictors subset $M$ and the minimal size $S$ have to be tuned. 

The interpretation of the link between the predictors and the event is not as easy in RSF as in (penalized) regression methods. To address this issue, RSF provide a quantification of this association, also known as variable importance (VIMP). For a given predictor $p$, $VIMP^{(p)}$ measures the mean (over all trees in the forest) increase of a tree error on its associated OOB sample, after randomly permuting the values of the $p^{th}$ predictor in the OOB sample. Large VIMP values indicate variables with prediction ability while null (or even negative) VIMP values indicate variables that could be removed from the prediction tool.

Using \texttt{rfsrc}() function (from \texttt{randomForestSRC} R package), three RSF sub-methods were considered that differed according to $M$ and $S$ parameter tuning: default software parameters $M=$ square root of the number of predictors, $S=15$ (called \textit{RSF-Default}), $M$ and $S$ that minimize the OOB error (called \textit{RSF-Optimize}) or $M$ and $S$ optimized plus a variable selection using the VIMP statistic (called \textit{RSF-SelectVar}).

\subsection{Step 4. Predicted individual probability of event}

The estimator of individual probability of event $\widehat{\pi}_{\star}^{(m)}(t_{LM},t_{Hor})$ for a new patient $\star$ becomes: 

\begin{itemize} 
\item For Cox, penalized-Cox and sPLS-DR models:
\begin{equation}
\widehat{\pi}_{\star}^{(m)}(t_{LM},t_{Hor}) = 1 - \exp{\left(- \widehat{\Lambda}_0(t_{Hor}) \exp{( \widehat{\mathcal{P}}_\star )}\right)} \label{eq:proba_indiv_phm}
\end{equation}

with $\widehat{\Lambda}_0(.)$ the Nelson-Aalen estimator, and $\widehat{\mathcal{P}}_\star$ the predicted linear predictor directly obtained from $\Gamma_\star$ and $X_\star$ for Cox and Penalized-Cox models, or from the $C$ components $f_c(\Gamma_\star, X_\star)$ ($c=1,...,C$) for sPLS-DR.

\item For RSF:
\begin{eqnarray}
\widehat{\pi}_{\star}^{(m)}(t_{LM},t_{Hor}) = 1 - \exp{\left(- \frac{1}{B} \sum_{b=1}^B \widehat{\Lambda}_\star^b(t_{Hor})\right)}
\end{eqnarray} 

with $\widehat{\Lambda}_\star^b(t_{Hor})$ the Nelson-Aalen estimator in the leaf of tree $b$ containing individual $\star$.

\end{itemize}

\subsection{Predictive accuracy assessment}\label{sec:assess}

We assessed the predictive performances of the models using the time-dependent Area Under the ROC Curve (AUC) \cite{blanche_estimating_2013} defined as: 
\begin{align}
    AUC(t_{LM},t_{Hor}) = P \Big( & \pi_i(t_{LM},t_{Hor}) > \pi_j(t_{LM},t_{Hor}) \Big| D_i(t_{LM},t_{Hor}) = 1 , \notag \\ 
    & D_j(t_{LM},t_{Hor}) = 0, T_i > t_{LM}, T_j > t_{LM} \Big) \label{eq:AUC}
\end{align}

and time-dependent Brier score \cite{mogensen_evaluating_2012} defined as:
\begin{equation}
BS(t_{LM},t_{Hor}) = E\Big[ \left( D_i(t_{LM},t_{Hor}) - \pi(t_{LM},t_{Hor}) \right)^2 \Big| T > t_{LM} \Big] \label{eq:BS}
\end{equation}

where $D_i(t_{LM},t_{Hor})$ is the survival status at time $t_{LM} + t_{Hor}$. We used estimators of these quantities that specifically handle the censored nature of $D_i(t_{LM},t_{Hor})$ using inverse censoring probability weighting (see \cite{mogensen_evaluating_2012,blanche_quantifying_2015} for details).

\begin{figure}[H]
  \centering
  \includegraphics[scale=0.5]{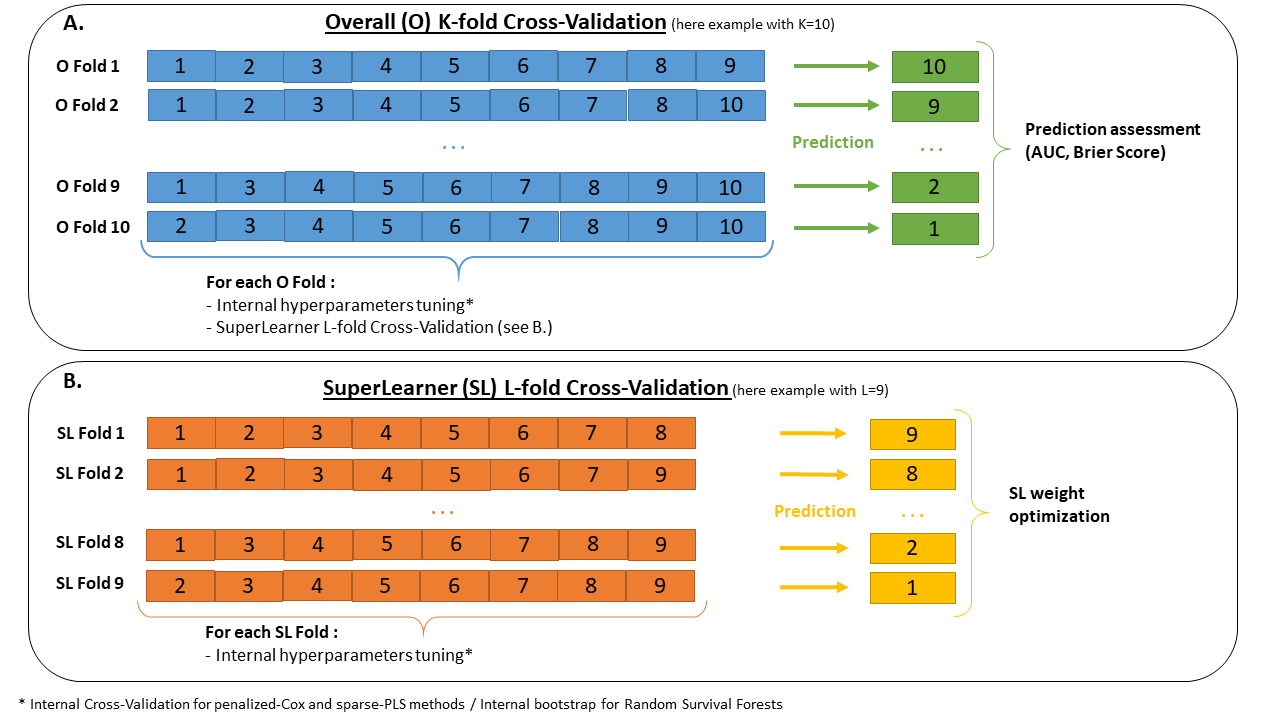}
  \caption{Multi-layer cross-validation framework: (A) Overall cross-validation to assess the predictive performances on independent samples, (B) Intermediate-layer cross-validation for the superlearner only performed on the learning sample to determine the weights. A final internal cross-validation (or Bootstrap for RSF) is done to tune each method.}
  \label{fig:SL_process}
\end{figure}

In the applications, predictive accuracy assessment was done using a cross-validation approach to ensure independence between the samples on which each predictive tool was learnt and the samples on which their predictive accuracy was assessed (Figure \ref{fig:SL_process}A). This induced a two-layer cross-validation since a cross-validation (or a bootstrap) was also performed within each training set to determine the method-specific hyperparameters.

\subsection{Combining the predictions into a single Super Learner}

Each survival prediction method $m$ ($m=1,...,M$) provides a different individual predicted probability $\widehat{\pi}^{(m)}_{\star}$ (equation \ref{eq:proba_indiv_estimator}). In some cases, one will prefer to select the best predictive tool and rely on it. In other cases, one can also choose to combine the predictive tools into a Super-Learner predictive tool \cite{van_der_laan_super_2007,golmakani_super_2020}. It consists in defining the final predicted probability as a weighted mean over the survival method-specific predictions:
\begin{equation}
\widehat{\Pi}_{\star} = \sum_{m=1}^M \omega_m \widehat{\pi}^{(m)}_{\star} \label{eq:SL}
\end{equation}

where the weights $\omega_m$ (defined in $[0,1]$ with $\sum_{m=1}^M \omega_m = 1$) are determined so that the Super-Learner predictive tool $\widehat{\Pi}$ minimizes a loss function. In our work, we chose to minimize the BS function defined in equation \eqref{eq:BS} by internal cross-validation. This lead to a three-layer cross-validation for the superlearner building and validation (see Figure \ref{fig:SL_process}B).

%%%%%%%%%%%%%
%% Results %%
%%
\section{Results}

\subsection{Performances of the methodology through a simulation study}

We contrasted the performances of the different survival prediction methods according to a serie of scenarios, based on Ishwaran \textit{et al.} \cite{ishwaran_random_2014}, in an extensive simulation study. Prediction tools were trained on $R = 250$ learning datasets and their predictive performances were compared on a unique external validation dataset. 

\subsubsection{Design}

The $R$ learning datasets and the validation dataset were generated according to the same design. They included $N = 500$ individuals at risk of the event at a landmark time $t_{LM}$ of 4 years. Up to landmark time, repeated information on 17 continuous biomarkers was generated according to linear mixed models, as described in equation \eqref{GLMM} with identity link.

For each biomarker, measurement times were randomly generated according to a $\mathcal{N}(0,0.15)$ around 5 theoretical visit times at -4, -3, -2, -1 and 0 years prior to $t_{LM}$. Different shapes of individual trajectory were considering depending on the biomarker, although all followed an individual polynomial function of time (see figure 1 in additional file). Summary characteristics of each error-free marker trajectory were computed (as defined in ``Methods" Section) leading to a total of 92 summaries statistics, stored in a vector $\Gamma_i^0$. An additional vector $X_i^0$ of 10 time-independent covariates was generated at the landmark time: 5 according to a standard normal distribution and 5 according to binomial distribution with success probability of 0.5.

The risk of event after the landmark time was defined according to a proportional hazard model with $\Gamma_i^0$ and $X_i^0$, and a Weibull distribution for the base hazard function, in order to not disadvantage the methods based on the Cox model.
Five different scenarios were built according to the number of summaries actually associated to the event (18 or 4 summaries) and the form of the dependence function: biomarkers summaries were entered into the linear predictor either linearly, linearly with interactions across biomarkers, or non-linearly with polynomial functions and binarization of summaries. Details on the generation model are given in section 1 of additional file.

The target prediction was the probability of event up to a horizon of 3 years. The predictive performances of all the survival methods were compared on the external dataset using the BS and AUC introduced in section \ref{sec:assess}, as well as the Mean Square Error of Prediction (MSEP), $MSEP = \frac{1}{N} \sum_{i=1}^N (\widehat{\pi}_i - \pi_i^0)^2$, which measures the average squared difference between the estimated probability $\widehat{\pi}_i$ and the true generated probability $\pi_i^0$ over all individuals.

\subsubsection{Results}

Predictive performances for scenarios with 18 summaries are summarized in Figure \ref{fig:resu_scenario21_42}. The same figure for scenarios with 4 summaries is given in Figure S2 of supplementary material.

When considering summaries entered linearly, the penalized-Cox provided the smallest BS and MSEP, and the highest AUC in both scenarios with 4 or 18 summaries associated with the event. When the relationships became increasingly complex (linear with interactions and non-linear), RSF provided better predictive performance than the other methods for both AUC, BS and MSEP regardless of the number of summaries considered. 

This simulation study highlights that the penalized-Cox model provides more accurate predictions in the case of simple relationships between the predictors and the event while RSF outperforms the others in the case of complex relationships (no matter how many summaries are considered). In contrast, classical Cox model was systematically outperformed by the other methods which points out the potential benefit of using advanced methods to predict the event in landmark approaches.

\subsection{Individual prediction of death in primary biliary cholangitis} \label{appli:PBC}

We first illustrated our method for predicting death among patients with primary biliary cholangitis (PBC). PBC is a chronic liver disease possibly leading to liver failure. For these patients, the only useful treatment is a liver transplantation \cite{kaplan_primary_1996}, and prediction of the risk of death can be useful in that context for patient stratification. We focused on the widely known PBC data from a clinical trial \cite{murtaugh_primary_1994} including repeated measures of 11 biomarkers (7 continuous and 4 binary), such as bilirubin value, albumin value or presence of hepatomegaly, and 3 additional demographic variables collected at the enrollment in the study (see Table S2 in supplementary material for the complete list). We aimed to predict the occurrence of death at horizon time $t_{Hor} = 3$ using information collected up to landmark time $t_{LM} = 4$ years on the $N = 225$ patients still at risk at $t_{LM}$ (see the flow chart Figure S3 in supplementary material).

After a normalization for continuous markers which did not follow a gaussian distribution using splines \cite{proust-lima_estimation_2017}, we modeled independently the markers according to generalized mixed models (see equation \ref{GLMM}) with natural splines on time measurements to capture potentially complex behavior over time \cite{perperoglou_review_2019} (See Section 3.1 in supplementary material for details on the models).

We used a 10-fold cross-validation to compute the predictive performances of the individual predicted probabilities. The distribution of the event times did not differ across folds (Figure S4 in supplementary material). For the superlearner, the optimal weights were determined in a second-layer 9-fold cross-validation. We repeated this process $R = 50$ times for all methods to assess the variability of the results across different cross-validation partitions.

Predictive performances are displayed in Figure \ref{fig:resu_pbc2}A. All the prediction tools provided satisfying predictive performance for both BS (from 0.076 to 0.089 in mean) and AUC (from 0.73 to 0.87 in mean). Nevertheless, we found that Cox models gave much worst indicators, especially for AUC (the only ones below 0.80 in mean), illustrating the limits of classical methods compared to machine learning methods that handle high dimension and correlation. In this application, the most discriminating and accurate predictions were given by the Cox model with Lasso penalty according to BS (0.076 in mean) and AUC (0.87 in mean). Results from the superlearner did not show substantial improvement in predictive performance. The weights of the superlearner indicated that it was mostly driven by penalized Cox models and RSF (Figure \ref{fig:resu_pbc2}B). 

For comparison, we also developed predictive tools based on (1) only baseline information for the 11 biomarkers and 3 covariates, (2) information on the 3 covariates and the trajectory of one biomarker over time (either serum bilirubin, albumin or platelets). The predictive tools based only on baseline information provided poorer cross-validation BS (32\% higher in mean over the methods) and AUC (8\% lower in mean over the methods) nicely illustrating the gain in updating the biomarker information over follow-up (Figure \ref{fig:pbc2_baseline_single}). The predictive performances were also worse when considering only repeated albumin or platelets with in mean 22\% and 37\% higher BS (1\% and 11\% lower AUC), respectively. In contrast, the predictive tools based on serum bilirubin (the main biomarker in PBC) provided similar performances as the multivariate predictive tool.

\subsection{Individual prediction of 5-years death at 80 and 85 years old in the general population}

In this second application, we aimed to predict the 5-year risk of death from any cause in the general older population at two different ages: 80 and 85 years old. We relied on the French prospective population-based aging cohort Paquid \cite{helmer_mortality_2001} which included 3777 individuals aged 65 years and older, and followed them up to more than 30 years with general health assessment every two to three years and continuous reporting of death. Beyond the individual quantification of the risk of death, our aim was to identify the main predictors of death and assess whether they differed according to age. The use of landmark models was perfectly adapted to this context with the definition of an age-specific prediction model. We chose to predict the 5-year risk of death from information on 9 markers of aging: depressive symptoms, 3 cognitive functions (general cognition, verbal fluency and executive function), functional dependency, incontinence, dyspnea, the live alone status, and polymedication as a global and easily collected marker of multimorbidity \cite{schneeweiss_performance_2001}. For each one, we focused on the trajectory over the last 5 years prior to the landmark age. In addition, we considered 18 other predictors including socio-demographic information (such as generation or gender) and medical history at the last visit prior to the landmark age (such as cardiovascular disease). Complete information on the markers and covariate definitions is given in Section 3.2 and Table S3 of supplementary material. The analysis was done on the samples of individuals still alive at $t_{LM} = 80$ and $t_{LM} = 85$, and with at least one measure for each of the predictors resulting in $N = 1561$ and $N = 1240$ subjects for $t_{LM} = 80$ and $t_{LM} = 85$, respectively (see flowchart Figure S6 in supplementary material). 

We used the exact same strategy as explained in the previous application for (i) modeling the trajectories of each marker except that time was the backward time (from -5 to 0 years) from landmark; (ii) computing the external probabilities with a 10-fold cross-validation and computing the superlearner with an internal 5-fold cross-validation. The event time distribution did not differ across folds (see Figures S7 and S8 in supplementary material). Note that due to the impossibility of using predictors with zero or near zero variance in sPLS-DR models, we removed from these models the following predictors: level of education, hearing, dementia, housing and dependency (ADL). RSF hyperparameters tuning (according to OOB error) is reported in supplementary material Figures S9 and S10.

Overall, the predictive performances of all the prediction models were very low with AUC ranging from 0.55 to 0.64 in mean and BS ranging from 0.123 to 0.135 in mean (see Figure S11A in supplementary material) showing the difficulty to accurately predict the age-specific risk of all-cause death in the general population. For both $t_{LM} = 80$ and $t_{LM} = 85$, RSF and the superlearner (which was mostly driven by the RSF (see Figures 11B and 12B in supplementary material) provided the lowest BS, whereas Cox with variable selection and penalized Cox models gave the highest AUC (0.66 in mean).

This application mainly aimed at identifying and contrasting the main age-specific predictors of death at 80 and 85 years old. Figure \ref{fig:var_rsfdefault} reports the VIMP from the optimized RSF (variables selected by the Lasso are shown in supplementary material Figure S14). The main predictors of 5-year death were mainly the trajectory of moderate functional dependency and polymedication both at 80 and 85 years old, dyspnea, gender and dementia at 80 years old as well as general self-assessment of health and severe dependency status at 85 years old. The predictors of 5-year death did not substantially differ between the two landmark times for RSF, except for dyspnea, general self-assessment of health and gender.

%%%%%%%%%%%%%%%%%
%% Discussion %%
%%
\section{Discussion}

We introduced in this paper an original methodology to compute individual dynamic predictions from a large number of time-dependent markers. We proposed to compute this prediction using a landmark approach combined with machine learning methods adapted to survival data. The idea was to incorporate a set of individual summaries of each marker trajectory (obtained in a preliminary longitudinal analysis) as well as other covariates in various prediction methods that could handle a large number of possibly correlated predictors, and complex associations. In addition to each prediction tool, we also proposed a superlearner, as a weighted mean of tool-specific predictions where weights were determined in an internal cross-validation to provide a minimal Brier Score. 

Through an extensive simulation study, we showed that regularized Cox models and RSF provided better cross-validated predictive performance over standard Cox model in different scenarios where there was a large number of markers and/or complex associations with the event. This was also observed in two real case applications: a clinical setting where death was predicted from monitored markers in primary biliary cholangitis, and in a setting where all-cause age-specific death was predicted in the general population from main markers of aging. 

Providing accurate predictions of health events that can exploit all the available individual information, even measured repeatedly over time, has become a major issue with the expansion of precise medicine. After the first proposals of dynamic predictions from repeated marker information \cite{proust-lima_development_2009,rizopoulos_dynamic_2011}, some authors have recently begun to tackle the problem of large dimension of longitudinal markers \cite{tanner_dynamic_2020,zhao_incorporating_2020,jiang_functional_2020}. In comparison with this recent literature, our method has the advantage of (i) considering any nature of markers with measurement error while other considered only continuous outcomes \cite{jiang_functional_2020}, (ii) proposing the use of many summaries from the biomarkers as individual posterior computation from the longitudinal model (compared for instance to \cite{tanner_dynamic_2020} who only include one or two summaries), (iii) exploiting the time-continuous information from survival data rather than discretized scale as in \cite{tanner_dynamic_2020}, and (iv) considering a vast variety of machine learning techniques as well as a superlearner rather than focusing only on one specific technique \cite{zhao_incorporating_2020}. 
Our methodology does not limit to the specific model and techniques described in the paper, it allows the use of any relevant method at each step. For example, we suggested to capture individual trajectories using generalized mixed models, but we also used functional principal component analysis \cite{yao_functional_2005} to characterize the individual variation of the trajectories using eigenfunctions leading to similar results (not shown here). We could also estimate the individual probability using other techniques such as deep learning \cite{katzman_deepsurv:_2018} or random forests based on pseudo-observations \cite{zhao_incorporating_2020}. Finally, although we considered for simplicity a single cause of event in this paper, our methodology could be extended to take into account several events through competing risks. For example, we could easily replace random survival forests by their extension that takes into account competing risks \cite{ishwaran_random_2014}. 

In our simulations and applications, we considered only a few dozens of markers repeatedly measured over time since this is already a challenging situation in individual dynamic prediction context where classical techniques are limited to a few markers. Yet, the method would also apply in a much higher dimensional context (e.g., with omics repeated data) or with a much larger amount of subjects. Indeed, thanks to the landmark approach, the prediction of the summary features for each biomarker and the prediction of the event are done in different steps, and each step-specific technique scales well in higher contexts (i.e., mixed model in large samples, and machine learning techniques in high dimensional context).

Our methodology is relevant for the prediction of an event from a landmark time that is common over subjects or for a small number of common landmark times as done in the application. In other settings where any landmark time should be considered, our methodology would need to be adapted as it currently involves as many prediction tools and the number of landmark times which would result in a considerable increase of computational burden. A possible solution might be to define the prediction tools as a continuous function of the landmarks, following the super landmark models idea \cite{houwelingen_dynamic_2012} but we leave such development for future research. 

%%%%%%%%%%%%%%%%%
%% Conclusions %%
%%
\section{Conclusions}

By extending the landmark approach to the large dimensional and repeated setting, our methodology addresses a current major issue in biomedical studies with a complete methodology that has the assets of being (i) easy to implement in standard software (R code is provided at https://github.com/anthonydevaux/hdlandmark) and (ii) generic as it can be used with any new machine learning technique adapted to survival data, any methodology to model repeated markers over time, any type of possible summary characteristics for the markers, and any number of markers.

\bibliographystyle{bmc-mathphys} % Style BST file (bmc-mathphys, vancouver, spbasic).
%\bibliography{bmc_article}      % Bibliography file (usually '*.bib' )
\bibliography{Biblio_Article1}

\section*{Figures}

\begin{figure}[H]
  \includegraphics{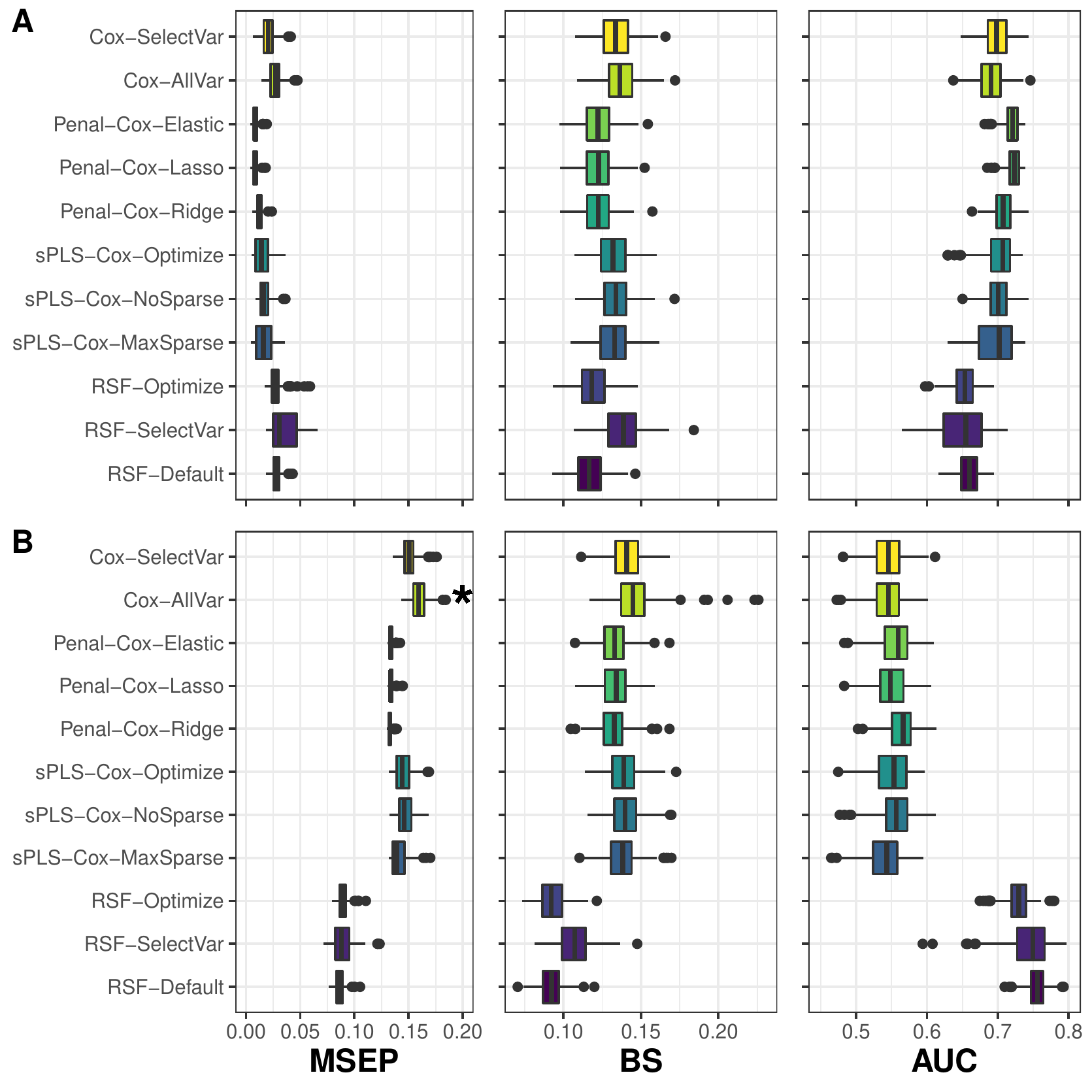}
  \caption{Simulation results over 250 replicates when considering 18 summaries associated to the event either assuming a linear form (figure A) or non-linear form (figure B). Methods are assessed using Mean Square Error of Prediction (MSEP), Brier Score (BS) and Area Under the ROC Curve (AUC). $(*)$ symbol indicates the presence of MSEP values above 0.2, but not displayed.}
  \label{fig:resu_scenario21_42}
\end{figure}

\begin{figure}[H]
  \includegraphics{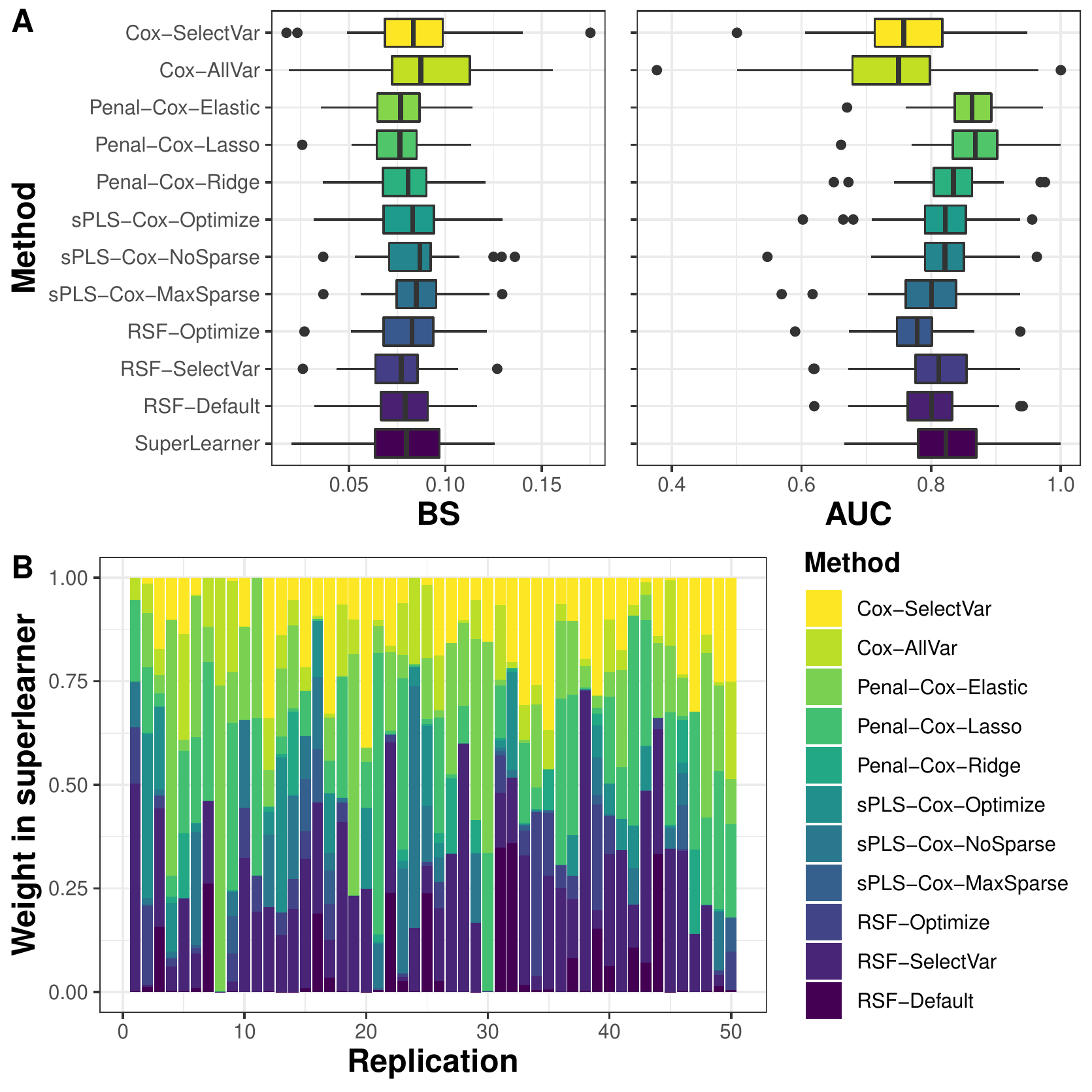}
  \caption{Assessment (figure A) and weights in superlearner (figure B) of 3-years death survival probability in primary biliary cholangitis patients using information collected up to 4 years over 50 replicates. Methods are assessed using Brier Score (BS) and Area Under the ROC Curve (AUC).}
  \label{fig:resu_pbc2}
\end{figure}

\begin{figure}[H]
  \centering
  \includegraphics{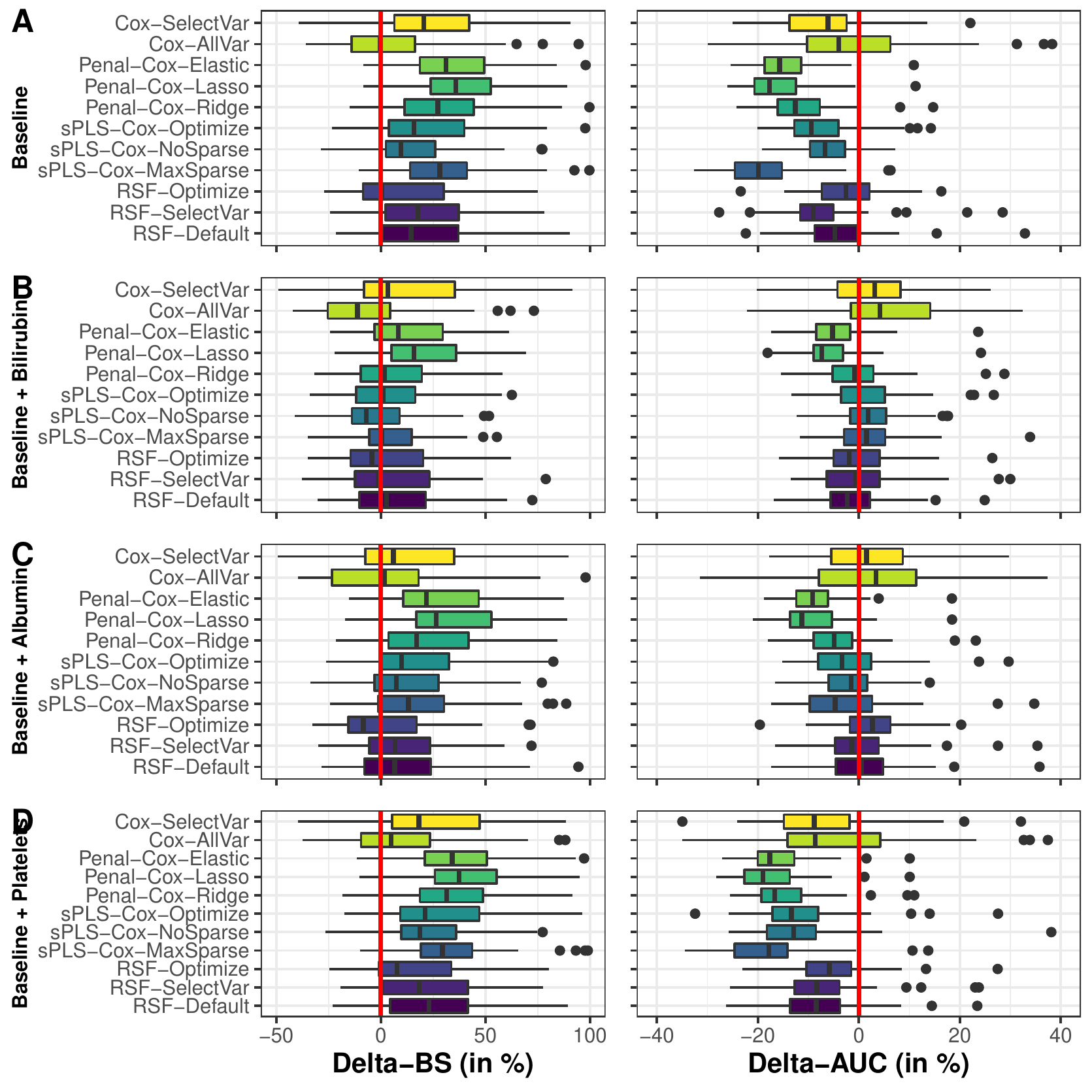}
  \caption{Assessment of 3-year survival probability in primary biliary cholangitis patients using baseline information on the 11 biomarkers and 3 covariates (figure A), baseline information and repeated measures collected up to 4 years of either serum bilirubin (figure B), albumin (figure C) or presence of platelets (figure D). The 10-fold cross-validation was replicated 50 times. The figure displays the difference (in percentage) of Brier Score (BS) and Area Under the ROC Curve (AUC) compared to the method using all the information with positive values for BS and negative values for AUC indicating a lower predictive accuracy.}
  \label{fig:pbc2_baseline_single}
\end{figure}

\begin{figure}[H]
  \includegraphics{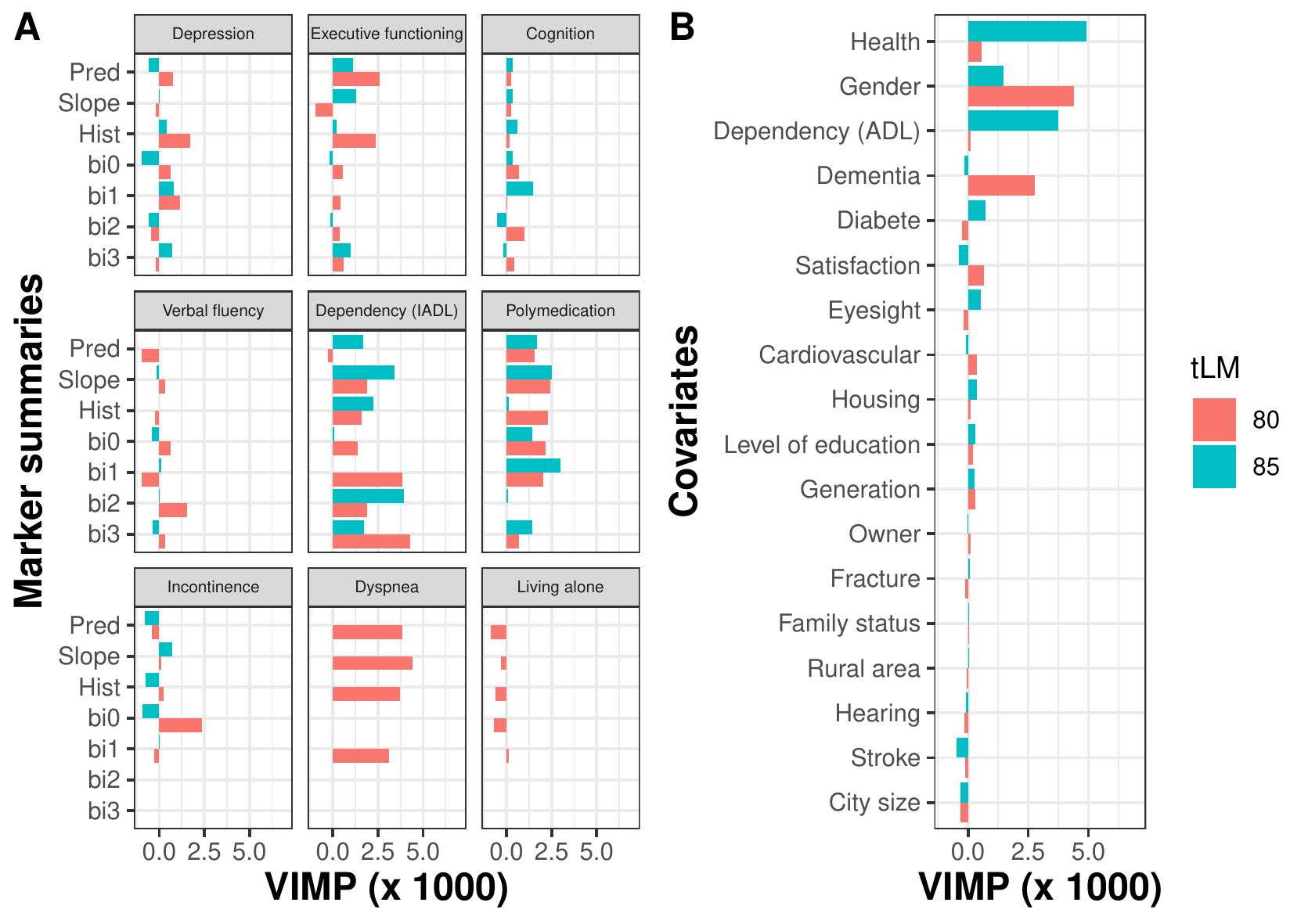}
  \caption{Variables associated with all-cause death in the \textit{RSF-Optimize} model at 80-year and 85-year landmark age. Are displayed the VIMP value for each marker summaries (figure A) and covariate (figure B). A large VIMP value indicates that the variable is predictive of the event.}
  \label{fig:var_rsfdefault}
\end{figure}

\end{document}